\documentclass[letterpaper, 10 pt, conference]{ieeeconf}  

\IEEEoverridecommandlockouts                              

\usepackage{graphicx}
\usepackage{comment}
\usepackage{float}
\usepackage{hyperref}
\usepackage{geometry}
\title{\LARGE \bf
Evaluating Robots Like Human Infants: A Case Study of \\ Learned Bipedal Locomotion
}

\author{Devin Crowley$^{1}$, Whitney G. Cole$^{2}$, Christina M. Hospodar$^{3}$, Ruiting Shen$^{4}$, Karen E. Adolph$^{5}$, and Alan Fern$^{6}$
\thanks{*This work was supported under NSF grant number 2321851.}
\thanks{$^{1}$Devin Crowley is with the Department of Electrical Engineering and Computer Sciences,
        Oregon State University
        {\tt\small crowleyd@oregonstate.edu}}
\thanks{$^{2}$Whitney G. Cole is with the Department of Psychology, New York University
        {\tt\small wgcole@nyu.edu}}
\thanks{$^{3}$Christina M. Hospodar is with the Department of Psychology, New York University
        {\tt\small christina.hospodar@nyu.edu}}
\thanks{$^{4}$Ruiting Shen is with the Department of Psychology, New York University
        {\tt\small rs8422@nyu.edu}}
\thanks{$^{5}$Karen E. Adolph is with the Department of Psychology, New York University
        {\tt\small karen.adolph@nyu.edu}}
\thanks{$^{6}$Alan Fern is with the Department of Electrical Engineering and Computer Sciences, Oregon State University
        {\tt\small alan.fern@oregonstate.edu}}
}

\begin{document}

\maketitle
\thispagestyle{empty}
\pagestyle{empty}

\begin{abstract}

Typically, learned robot controllers are trained via relatively unsystematic regimens and evaluated with coarse-grained outcome measures such as average cumulative reward. The typical approach is useful to compare learning algorithms but provides limited insight into the effects of different training regimens and little understanding about the richness and complexity of learned behaviors. Likewise, 
human infants and other animals are ``trained" via unsystematic regimens, but in contrast, developmental psychologists evaluate their performance in highly-controlled experiments with fine-grained measures such as success, speed of walking, and prospective adjustments. However, the study of learned behavior in human infants is limited by the practical constraints of training and testing babies. Here, we present a case study that applies methods from developmental psychology to study the learned behavior of the simulated bipedal robot Cassie. Following research on infant walking, we systematically designed reinforcement learning training regimens and tested the resulting controllers in simulated environments analogous to those used for babies---but without the practical constraints. Results reveal new insights into the behavioral impact of different training regimens and the development of Cassie's learned behaviors relative to infants who are learning to walk. This interdisciplinary baby-robot approach provides inspiration for future research designed to systematically test effects of training on the development of complex learned robot behaviors.

\end{abstract}


\begin{figure*}[!htb]
    \centering
    \includegraphics[width=\textwidth]{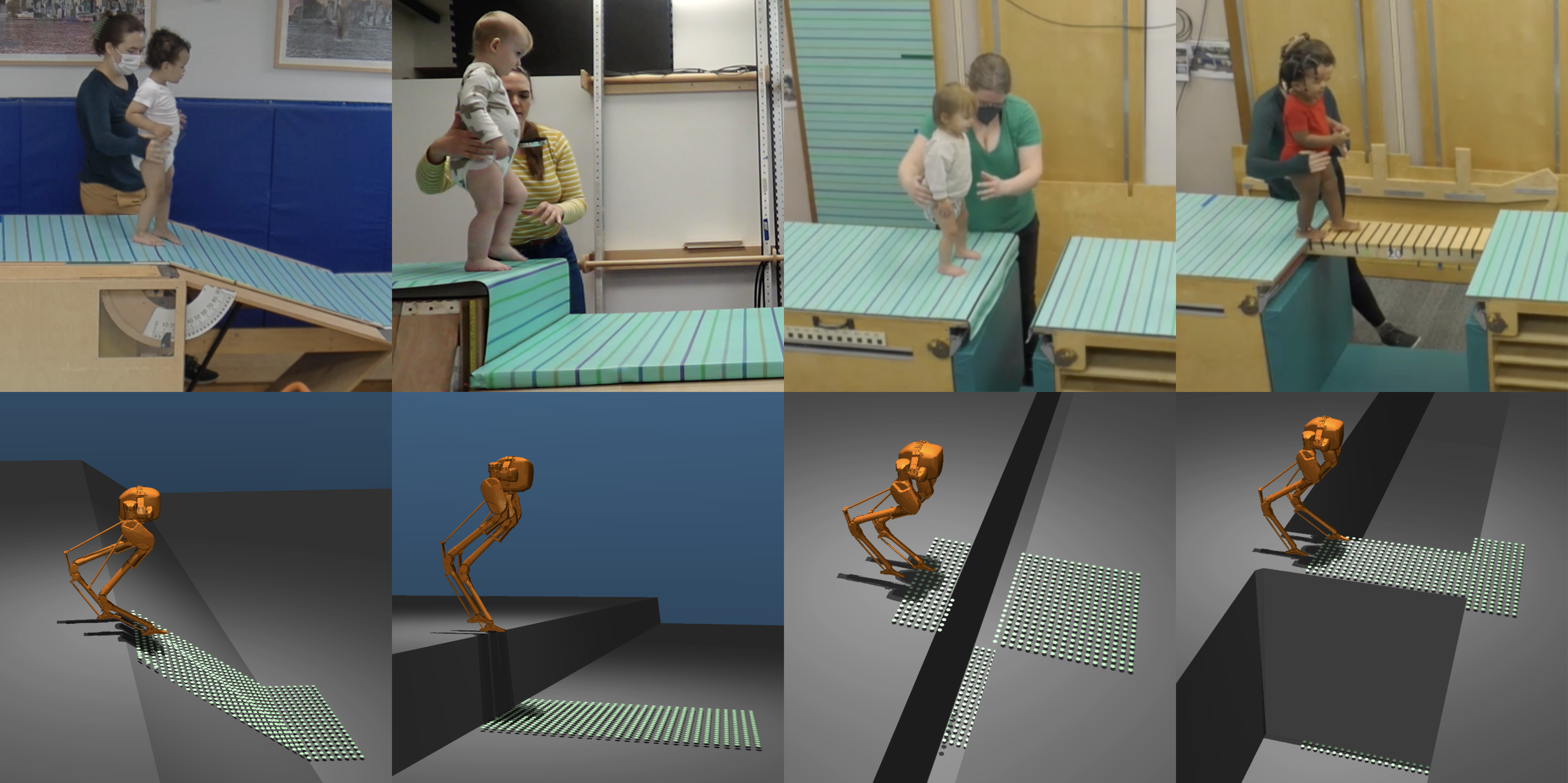}
    \caption{Test environments with infants in the real world and the bipedal robot Cassie in simulation. Left-to-right: slopes, drop-offs, gaps, and bridges.}
    \label{fig:human_robot_composite}
    \vspace{-5mm}
\end{figure*}

\vspace{-0.2em}
\section{Introduction}
\vspace{-0.2em}
\label{sec:introduction}

Consider a visually-guided bipedal robot trained in simulation to walk over challenging terrain with varied elevations and obstacles in the path. Typically, training regimens and test situations are not manipulated systematically, and performance evaluations report only crude metrics (e.g., average cumulative reward over multiple test environments). Albeit useful to roughly rank learned controllers, 
the typical approach to training, testing, and evaluation cannot reveal how different training regimens affect the complex details of learned behaviors like gait modifications while approaching and navigating varied terrain. For example, two different controllers with similar reward functions may nonetheless behave very differently when walking down a steep slope. One controller might leverage its vision and adjust its gait prospectively while approaching the slope, whereas the other controller may adjust reactively after stepping on the slope. 

Inspired by developmental research with human infants, we advocate for a more systematic approach to training and testing, and a more detailed evaluation of the learned behaviors. To promote this approach, this paper presents a case study that adapts developmental research on infant locomotion to learned, visually-guided locomotion controllers for the Cassie bipedal robot. In particular, we used simulated test environments with the experimental apparatuses from infant studies (slopes, drop-offs, gaps, and bridges) and conducted similar experiments and analyses \cite{adolph2019annreview, adolph2021visualcliff} (see Fig. \ref{fig:human_robot_composite}). 

Testing in simulation removes the constraints posed by real-world robot experiments, allowing fine-grained comparisons of different training regimens. Simulation-based work is often accompanied by a sim-to-real transfer method to produce a controller that functions in the real world. However, addressing the reality gap is orthogonal to the investigations of this work. We test in simulation without transfer to the real world because our objective is to study the development of locomotion capabilities, not in producing a working real-world controller.

This robot evaluation approach can advance behavioral research with robots by guiding robot training regimens and informing decisions about which controller works best in real-world operating environments. 
In addition, this approach may also inform behavioral research with infants by revealing the benefits and constraints of reinforcement learning (RL) under various training regimens---regimens that would be unethical or impractical for human babies.

\vspace{-0.2em}
\section{Background}
\vspace{-0.2em}
\label{sec:background}

Our case study follows prior work on learning locomotion controllers for the Cassie robot and analyzes behavior using established methods from research with walking infants.

\vspace{-0.4em}
\subsection{Learning in the Development of Infant Locomotion}
\vspace{-0.4em}

How do babies learn to navigate varied terrain such as steep slopes, high drop-offs, wide gaps, and narrow bridges, as illustrated in  Fig. \ref{fig:human_robot_composite}? Infants learn to walk amidst continual changes in their environments and skills. Features of the environment are variable (ground surfaces can be flat or sloping, rigid or deformable, high-traction or slippery; the path can be clear or cluttered with obstacles and elevations) and infants' walking skill improves dramatically over the first several months after walk onset \cite{adolph2019annreview, adolph2018tics, hospodar2024wires}. During natural locomotion, infants' walking paths are curved, not straight; their steps are omnidirectional, not forward; and their locomotion is intermittent rather than continuous \cite{cole2016bouts, lee2018cost, hospodar2021practice, adolph2012learntowalk}. Infants' everyday life creates a natural training regimen---with toys scattered on the ground, laundry heaps in the corner, furniture obstructing the path, and so on. For infants, learning to walk means learning to modify gait from step to step to navigate a varied environment. 

Developmental researchers test novice and experienced walking infants on challenging terrain, such as slopes, drop-offs, gaps, and bridges of varying difficulties, because the obstacles are novel (no baby encounters steep slopes, narrow bridges, etc. during everyday life). Thus, researchers can test generalization from everyday experience to novel situations based on whether babies modify their gait prospectively while approaching the obstacle or reactively after stepping onto the obstacle. In all cases, the best solution is to modify gait prospectively on approach. Reactive adjustments while traversing the obstacles are less optimal because gravity takes over and gait modifications entail fighting to keep the body over the moving base of support. The strongest evidence for prospective control is changes in foot placement, step length, and walking speed before stepping onto the obstacle---rather than modifying gait after stepping onto the obstacle.

Prior work suggests the ability to modify gait develops with walking experience \cite{adolph2019annreview,adolph2018tics, hospodar2024wires}. Novice infants walk blithely over the brink of impossibly steep slopes, high drop-offs, wide gaps, and narrow bridges (requiring rescue from an experimenter). After several months of everyday walking experience---notably, with no prior experience on the test obstacles---infants modify their gait prospectively. For every obstacle, experienced infants slow down and shorten their steps during approach. For slopes, their initial steps on the slope are short and slow, they widen their base of support, and brake forward momentum from step to step \cite{adolph1997learning, gill2009change}. For drop-offs, they place their stance foot close to the brink so that their moving foot can stretch down to the bottom of the precipice all the while keeping their body upright as it drops vertically onto their moving foot \cite{kretch2013cliff}. For gaps, they place their stance foot close to the brink of the gap, increase step length with the moving foot, and place their moving foot close to the far side of the gap. For bridges, infants place their body at the near side of the bridge, and take short, slow, narrow steps \cite{kretch2013bridge, kretch2017organization}.

Developmental researchers hypothesize that infants learn to modify gait prospectively as they accumulate everyday experience navigating an ever-changing environment. Practice doing the same thing repeatedly (e.g., stepping on a treadmill) and training to the test (e.g., repeated practice walking down slopes) do not contribute to the development of prospective gait modifications \cite{gill2009change}. However, developmental researchers cannot control infants’ natural, everyday training regimens and must accommodate practical limitations for testing infants. Babies can complete only a few dozen trials before becoming tired or fussy, and often learn gait modifications During testing. Such limitations do not exist for simulated robots. 
Thus, simulated robots provide a powerful, highly-controlled platform for testing effects of experience on motor learning, with the potential to serve as a surrogate for studying how infants learn to walk over varied terrain.

\vspace{-0.4em}
\subsection{RL for Bipedal Locomotion}
\vspace{-0.4em}

\begin{figure*}[!htb]
    \centering
    \includegraphics[width=\textwidth]{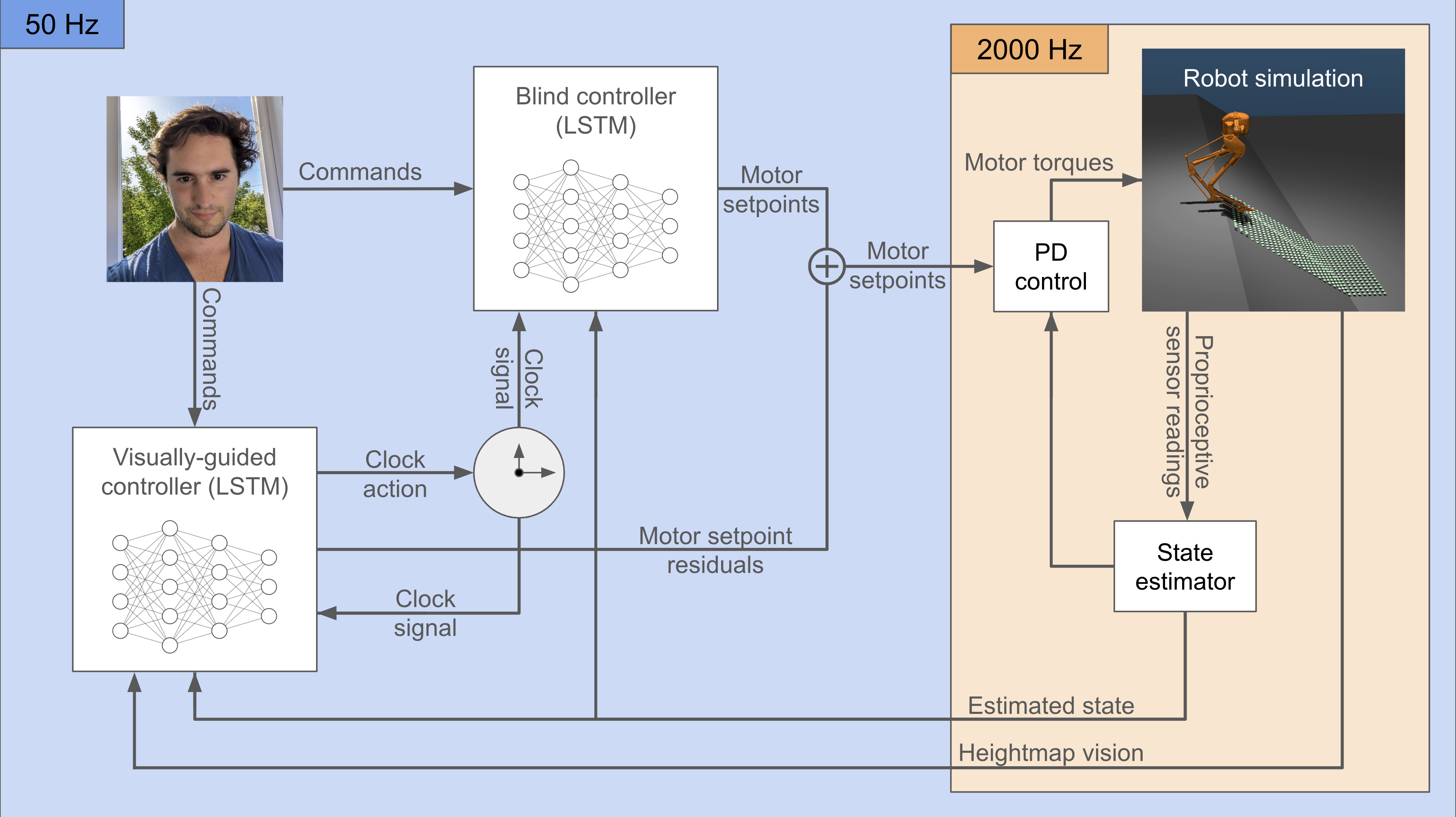}
    \caption{Controller schematic. The controller consists of a visually-guided component, trained on varied terrain, which modulates the output of a blind component trained on only flat ground.}
    \label{fig:rl_schematic}
\end{figure*}

\textbf{Controller Architecture}
We tested the behavior of controllers for the Cassie bipedal robot trained with RL using a learning framework from prior work \cite{duan2023learning}, lightly adapted for purely simulation-based experimentation. This framework layers a visually-guided residual controller on top of the outputs of a frozen blind controller trained only on flat terrain. The controllers are represented as recurrent long short-term memory (LSTM) neural networks. Their outputs are proportional derivative (PD) control targets, set at 50 Hz for 10 actuated joints centered around a standing pose. A static PD controller uses these targets to set the motor torques at 2000 Hz. A schematic of this learning framework is shown in Fig. \ref{fig:rl_schematic}. Training a slower controller to set targets for a faster PD controller is commonly used in RL for locomotion \cite{peng2017learning, xie2018feedback, tan2018AgileLocomotion, hwangbo2019learning, TsounisDeepGait} and is a strong precedent for the Cassie platform \cite{duan2023learning, siekmann2020learning, siekmann2020simtoreal, siekmann2021stairs}.

The input to the blind controller includes: (1) a 35-dimensional vector containing the positions and velocities of all joints, plus the pelvis orientation and rotational velocity; (2) a clock signal that dictates the cadence of the footsteps; (3) gait parameters that modulate the clock signal; and (4) commands for forward speed, lateral speed, and turn rate. A grid of ground-truth terrain heights taken from the simulated terrain serves as additional input to the visually-guided controller. The grid is a 1m wide by 1.5m long rectangle of 20 by 30 values respectively in front of Cassie.

In addition to PD target residuals, the visually-guided controller outputs the clock progression speed and the phase-offset between the feet. This allows the controller to modulate the frequency of footsteps and adjust the left-right cadence. In principle, this means the controller can learn to produce asymmetric 2-beat patterns like skipping rather than  performing only the basic, symmetrical, left-right gait pattern. 
Both blind and visually-guided controllers use the same reward function. It encourages adherence to the commands, minimized motor torque, and footsteps in accordance with the clock signal.
See \cite{duan2023learning} for further details.

\textbf{Simulation Training}
The controllers are trained using the actor-critic proximal policy optimization (PPO) algorithm with gradient clipping, a standard model-free RL algorithm \cite{schulman2017proximal}. The simulator used for both training and testing is the MuJoCo physics engine \cite{todorov2012mujoco} using a model of the Cassie robot. Because we examined behavior only in simulation, we did not use dynamics randomization to aid in sim-to-real transfer as in prior work with Cassie \cite{duan2023learning, siekmann2020learning, siekmann2020simtoreal, siekmann2021stairs, yu2022dynamic, dao2023simtoreal, crowley2023gaits}.

\vspace{-0.2em}
\section{Experimental Setup}
\vspace{-0.2em}
\label{sec:experimental setup}

Our analyses evaluate the performance and behavior of several visually-guided controllers, differing in the distribution of training terrains. All controllers output residual PD targets added to the outputs of the same pre-trained blind controller described in Section \ref{sec:background}.

\vspace{-0.4em}
\subsection{Training Regimens}
\begin{figure*}[!htb]
    \centering
    \includegraphics[width=\textwidth]{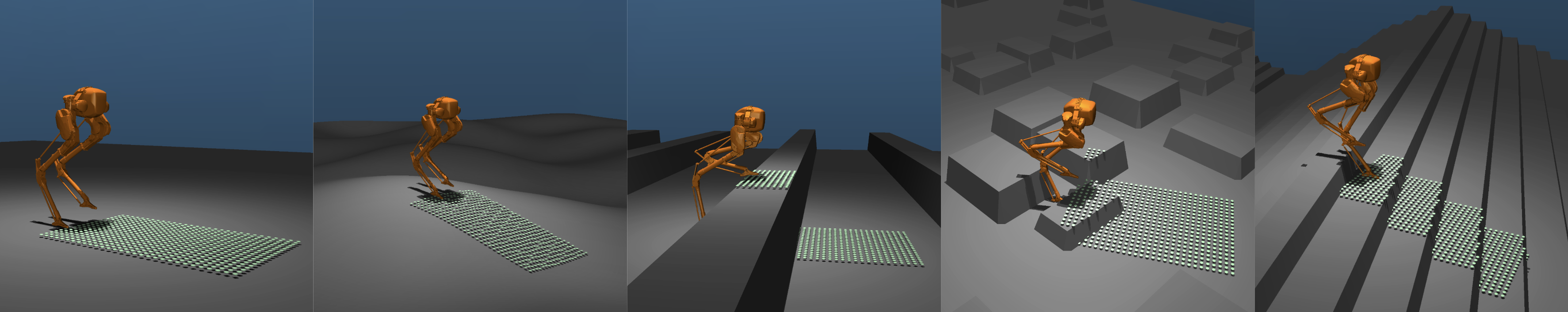}
    \caption{Standard terrains. Left-to-right: flat, hills, ridges, blocks, and stairs.}
    \label{fig:icra_terrains}
\end{figure*}

\vspace{-0.4em}
We trained Cassie on \textit{standard}, \textit{multi-test obstacle}, \textit{combined standard and multi-test}, and \textit{single-test obstacle} terrains.  The standard training terrains replicate prior robotics work \cite{duan2023learning} on which our learning framework is based: flat, hills, ridges, blocks, and stairs, shown in Fig. \ref{fig:icra_terrains}. The multi-test obstacle terrains are recreations of test apparatuses used in experiments with infants: slopes, drop-offs, gaps, and bridges, shown in Fig. \ref{fig:human_robot_composite} for infants in the real world and Cassie in simulation. 

Table \ref{tab:curricula} shows the frequency of each terrain for the standard, multi-test obstacle, combined standard and multi-test, and single-test obstacle training regimens. Cassie received 20k iterations for the single-test obstacle regimen and 110k iterations for each of the others. The standard regimen uses a range of commands for forward speed, lateral speed, and turn rate. The test-obstacle regimens use only one command: straight forward at $0.8\frac{m}{s}$.

\begingroup
\renewcommand*{\arraystretch}{1.1}
\begin{table*}[tb]
    \caption{Terrain frequencies by training regimen}
    \begin{center}
        \begin{tabular}{|c||c|c|c|c|c||c|c|c|c|}
            \hline
            \rule{0pt}{0.9\normalbaselineskip} 
            Terrain set & 
            \multicolumn{5}{|c||}{Standard} & 
            \multicolumn{4}{|c|}{Test Obstacles}\\
            \hline \hline
            \rule{0pt}{0.9\normalbaselineskip} 
            Terrain & Flat & Hills & Ridges & Blocks & Stairs & Slopes & Drop-offs& Bridges & Gaps\\
            \hline \hline
            \rule{0pt}{0.9\normalbaselineskip} 
            Standard & 3\% & 7\% & 20\% & 35\% & 35\% & 0\% & 0\% & 0\% & 0\%\\
            Multi-test & 0\% & 0\% & 0\% & 0\% & 0\% & 25\% & 25\% & 25\% & 25\%\\
            Combined & 1.5\% & 3.5\% & 10\% & 18.5\% & 18.5\% & 12.5\% & 12.5\% & 12.5\% & 12.5\%\\
            Single-test & 0\% & 0\% & 0\% & 0\% & 0\% & 100\% & 0\% & 0\% & 0\%\\
            Single-test & 0\% & 0\% & 0\% & 0\% & 0\% & 0\% & 100\% & 0\% & 0\%\\
            Single-test & 0\% & 0\% & 0\% & 0\% & 0\% & 0\% & 0\% & 100\% & 0\%\\
            Single-test & 0\% & 0\% & 0\% & 0\% & 0\% & 0\% & 0\% & 0\% & 100\%\\
            \hline
        \end{tabular}
    \end{center}
\vspace{-5mm}
\label{tab:curricula}
\end{table*}
\endgroup

\vspace{-0.4em}
\subsection{Testing Setup}
\vspace{-0.4em}
Our controller evaluation mirrored tests with infants, so we used only the four obstacle terrains. Thus, terrains used in the obstacle training regimens are identical to those used in testing.  

Cassie began each test trial facing the obstacle (slope, drop-off, gap, or bridge) at a distance of 3-3.5m, with a lateral offset of 0-0.25m. Each obstacle had a difficulty parameter, ranging from 0-1, randomized in training, that linearly adjusts the relevant property for each terrain. Cassie received 50 trials at each of 101 difficulty levels, for a total of 5050 trials per obstacle. The downward angle of slopes (1.5m long) ranged from 0-90\textdegree, the step-down height of drop-offs ranged from 0-1.5m, bridge width (1.5m long) ranged from 0.02-1.02m, and gap width ranged from 0-1m.

\vspace{-0.2em}
\section{Behavioral Analysis Results}
\vspace{-0.2em}
\label{sec:behavioral analysis results}

\begin{figure*}[!htb]
    \centering
    \includegraphics[height=0.925\textheight]{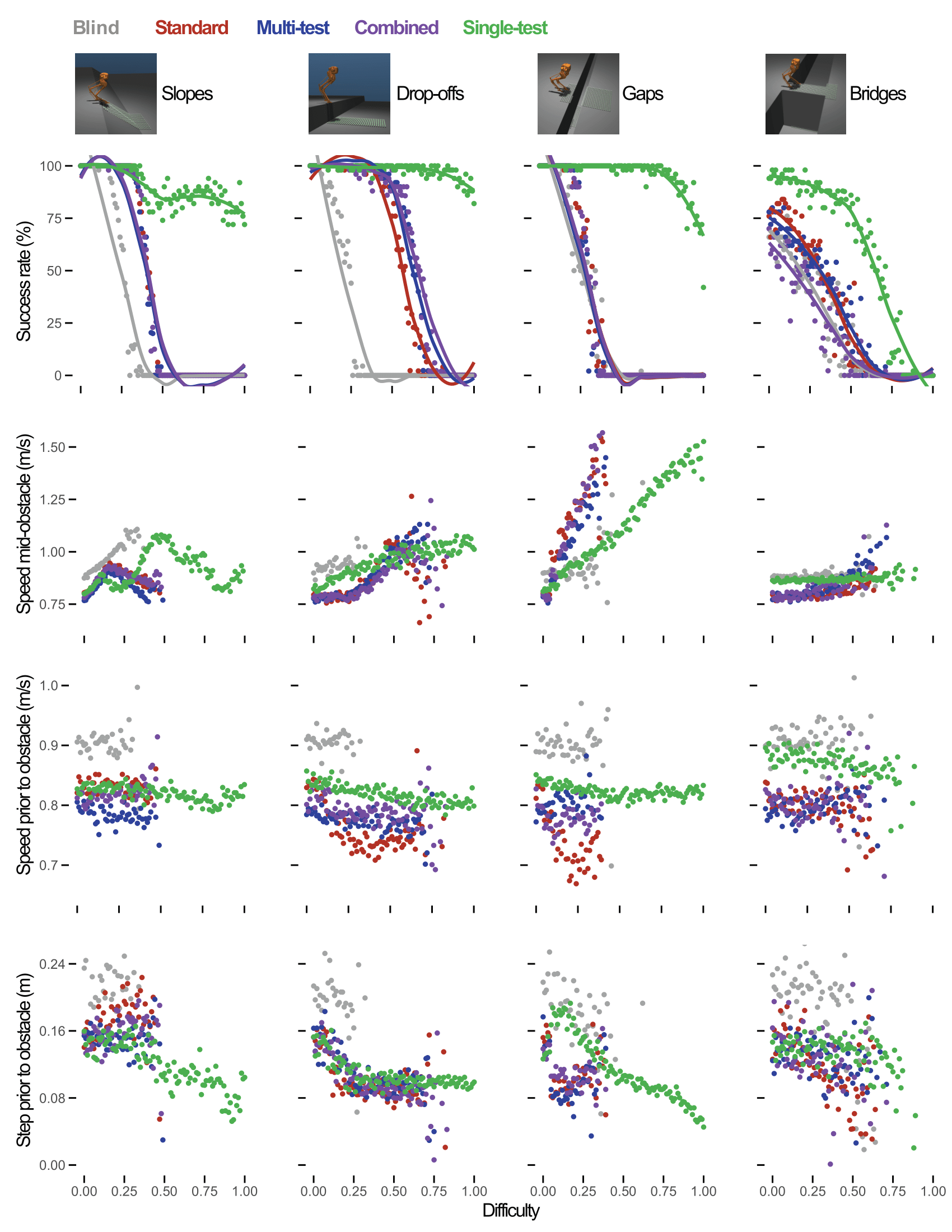}
    \caption{Evaluation on slopes, drop-offs, gaps, and bridges. Each column shows behavioral results for each test obstacle across continual, systematic increase in difficulty. Curves show blind, standard, multi-test obstacle, combined standard and multi-test obstacle, and single-test obstacle training regimens. Top row: Success rates for navigating the obstacles. Second row: Average speed of walking on or over the obstacle. Third row: Average speed of last two steps prior to the obstacle. Bottom row: Placement of last step relative to the edge of the obstacle.}
    \label{fig:megafigure}
    \vspace{-2em}
\end{figure*}

\vspace{-0.4em}
\subsection{Success Rate}
\vspace{-0.4em}
Training improved Cassie's success at navigating obstacles, but performance varied depending on the obstacle (Fig. \ref{fig:megafigure}, top row). Notably, the single-test obstacle regimens (green curves) ensured greater success on every obstacle compared with the other regimens. On slopes and drop-offs, every training regimen improved success relative to the blind controller (gray curves), and on the drop-offs, the multi-test and combined regimens improved performance relative to the standard regimen (red curve). However, on gaps and bridges, results were mixed. On gaps, the standard, multi-test, and combined regimens performed equivalently to the blind controller. And on bridges---even at the lowest difficulty level---blind, standard, multi-test, and combined regimens produced success rates of $<78\%$. Yet, training to the test in the single-test obstacle regimen demonstrates that the gap and bridge obstacles were learnable.

\vspace{-0.4em}
\subsection{Gait Modifications Mid-Obstacle}
\vspace{-0.4em}
Successful walking on more difficult obstacles was achieved in part by modifying gait after stepping on or over the obstacle (Fig. \ref{fig:megafigure}, second row)---that is, in the multiple steps on the slope or bridge and the single step to cross the drop-off or gap. On slopes, for example, the speed of the blind controller (gray curve) increased with difficulty as gravity pulled Cassie down steeper slopes; speed peaked at $\sim$30\% difficulty and Cassie failed thereafter. The standard, multi-test, and combined training controllers also increased speed with difficulty, but speed peaked at  $\sim$22\% difficulty---before the peak of the blind controller---then decreased on steeper slopes as Cassie began using a braking strategy, resulting in success on steeper slopes than the blind controller could manage. The single-test training controller initially peaked even earlier at  $\sim$11\% difficulty, before implementing a braking strategy. The second peak for the single-test controller resulted from Cassie slipping down the slope, and speed decreased as Cassie began jumping down the slope at even more difficult increments.

On drop-offs and gaps, speed increased with difficulty. For drop-offs, increased speed likely reflects effects of gravity pulling the body down, but for gaps, it likely reflects Cassie launching its body to span larger gaps. Consistent with the poor success rate on bridges, Cassie increased speed on narrower bridges---making narrower bridges more challenging.

Although mid-obstacle gait modifications indicate that Cassie modified its gait to cope with more difficult obstacles, we cannot definitively categorize such adjustments as prospective. Increased speed to launch over a gap is produced before crossing (i.e., prospective), but decreased speed after stepping onto the slope could be in reaction to feeling the slant (i.e., reactive), and increased speed on the drop-off may be entirely out of Cassie's control (i.e., neither prospective nor reactive). The best test of planning, therefore, is gait modifications \textit{prior} to encountering the obstacle.

\vspace{-0.4em}
\subsection{Gait Modifications Prior to the Obstacle}
\vspace{-0.4em}
\label{sec:gait modifications prior to the obstacle}
Cassie did not show compelling evidence of prospective speed adjustments prior to obstacles (Fig. \ref{fig:megafigure}, row 3). Speed and step length in the preceding two steps were constant or decreased only slightly across difficulty levels (e.g., from $\sim$0.83 m/s at 0 difficulty to $\sim$0.76 m/s at 50\% difficulty).

Cassie did, however, show evidence of prospective gait modifications based on foot placement (Fig. \ref{fig:megafigure}, row 4). 
Cassie placed its last step prior to the obstacle closer to the edge as difficulty increased for the single-test regimen on slopes, drop-offs, and gaps, and for the standard, multi-test, and combined regimens on drop-offs and gaps. Placing the foot close to the edge is crucial, especially for drop-offs and gaps because it shortens the size of the step needed to cross. For drop-offs, the standard, multi-test, and combined regimen’s last step landed $\sim$0.15m from the edge at difficulty 0, but dropped to $\sim$0.08m on drop-offs at difficulty $\sim$25-100\%.

\vspace{-0.2em}
\section{Discussion}
\vspace{-0.2em}
\label{sec:discussion}
We applied experimental methods from research with human infants to the simulated bipedal robot Cassie. Systematic manipulation of Cassie's training regimens and tests of performance outcomes revealed differential effects on learning based on a detailed characterization of behavior. Training specifically to the test (single-test obstacle regimen) resulted in higher success rates and more prospective gait modifications than training on a variety of non-test terrains (standard regimen), a variety of test obstacles (multi-test obstacle regimen), or a combination of non-test and test obstacles (combined regimen).

\vspace{-0.4em}
\subsection{Effects of Training} 
\vspace{-0.4em}
\label{sec:effects of training}

Cassie showed markedly superior performance on each test obstacle when trained exclusively on that obstacle compared to the other training regimens. This indicates limited generalizability between these terrains, despite qualitative similarities. Hills are akin to slopes; ridges, blocks, and stairs are akin to drop-offs. Even the test obstacles approximate each other at some levels. The steepest slope is identical to the highest drop-off, and the widest gap approximates the gap around the bridge. The bridge obstacle is the most unique, requiring Cassie to not only handle the terrain in front of it, but to adjust navigation and guide foot placement to keep from falling off. We see from the poor single-test performance on bridges that it is also the most difficult. This may be explained somewhat by the reward function encouraging an uncompromising heading.

Interestingly, the success rates are scarcely improved from the standard regimen to the multi-test and combined regimens where Cassie is exposed to the test obstacles. The single-test controllers were trained for 20k iterations on the obstacle they would be evaluated on, whereas the controllers trained on the standard, multi-test obstacle, and combined standard and multi-test regimens were trained for 110k iterations spread across various terrains. The multi-test regimen therefore received 27.5k iterations on each test obstacle, but still shows inferior performance to single-test. This indicates that the variety of experience is hindering Cassie's ability to learn the best strategies for each obstacle or possibly that longer training runs or larger models are required.

This interpretation is corroborated by the speeds mid-obstacle in Fig. \ref{fig:megafigure}, row 2. The single-test (green) curves stand apart from the others, indicating a different strategy was learned. This is most easily seen on slopes (column 1), where a braking behavior is adopted earliest (at a lower difficulty) by the single-test controller. Even more telling is the shift where it speeds up again, indicating a change in strategy not seen by the other controllers.

As discussed in Section \ref{sec:gait modifications prior to the obstacle}, Cassie does not modulate its speed preceding an obstacle, but it does adjust foot placement, as illustrated in the last two rows of Fig. \ref{fig:megafigure}. The more the speed deviates from the commanded speed the more reward is lost, so an inflexible reward function may account for the consistent speeds. However, the intentional foot placement is most clearly seen on the gap obstacle in the bottom row of Fig. \ref{fig:megafigure}. Curiously, the single-test step lengths are larger and more similar to the blind controller's, whereas the other regimens produce shorter preceding footsteps, possibly indicating a reluctance to step off the ledge. This combined with their low success rate indicates that they haven't discovered an effective strategy (stepping across the gap) and are falling back on optimized failure modes.

\vspace{-0.4em}
\subsection{Differences Between Babies and Robots}
\vspace{-0.4em}

Cassie demonstrated motor skills more advanced than any infant---jumping down high drop-offs and recovering from near-catastrophic falls. But we tested simulated Cassie. Real Cassie---with a physical body---would have required repairs after such feats. Babies also must deal with the consequences of errors, but cannot be taken to the shop for repairs. Instead, their body is built to cope with frequent errors in learning to walk: infants are small, low to the ground, and move slowly, decreasing impact forces produced by a fall \cite{han2021impact}.

Moreover, every baby learns things Cassie did not learn: Infants use a wide range of prospective gait modifications, invent and use alternative strategies for obstacles where modifications are insufficient, and avoid crossing impossible obstacles \cite{adolph2019annreview, adolph2018tics, hospodar2024wires}. Most critical, babies generalize learning. No infant experiences single-test training. To acquire behavioral flexibility and prospective control of locomotion, infants must generalize from everyday experiences. In this regard, any 18-month-old can run circles around Cassie. Infants' greater flexibility and adaptability may result from more powerful learning mechanisms than pure RL. One hypothesis is that infants are ``learning to learn". That is, they learn to generate and gather the relevant information and use it to guide their actions from moment to moment.

\vspace{-0.2em}
\section{Limitations and Future Work}
\vspace{-0.2em}
\label{sec:limitations}
Our case study tested controller behavior only at a single point in training with a given regimen. Future work should investigate the development of learned behaviors at varied points in training to understand the trajectory of learning.

In contrast to infants, Cassie did not prospectively modify speed while approaching obstacles. Apparently, infants' natural training regimen teaches them to prospectively modify their speed to better cope with obstacles even as they are learning to walk. Decreased speed is useful for walking down steep slopes, high drop-offs, and narrow bridges, and increased speed is useful to leap over wide gaps. However, Cassie received an explicit, fixed speed command, and was encouraged to abide by it in the reward function. Thus, speed adjustments must yield greater improvements in the reward than the reward lost due to the speed error, or else those behaviors won't be learned. Our fixed speed command may have precluded Cassie's discovery of speed adjustment. To improve robot learning and to better understand infant walking, future work should consider relaxing the speed command to give the controller greater flexibility to modify gait and adopt novel strategies.


\vspace{-0.2em}
\section{Conclusions}
\vspace{-0.2em}
\label{sec:conclusion}
The simulated bipedal robot Cassie learned to modify its gait via precise foot placement just-prior to the obstacle and modifying speed while walking on or stepping over the obstacle. However, systematic training and testing based on methods from developmental research with human infants revealed that every training regimen produced more limited generalization and less adaptive behavioral modifications than expected, but a greater ability to jump, launch, and recover balance after large-amplitude movements and near falls. Likely, babies beat bots because they are ``learning to learn" rather than responding solely to rewards. 


\addtolength{\textheight}{-12cm}   




\bibliographystyle{IEEEtran}
\vspace{-0.4em}
\bibliography{IEEEabrv, mybib}  
\vspace{-0.4em}

\end{document}